\documentclass[runningheads]{llncs}
\usepackage{eccv}
\usepackage{eccvabbrv}
\usepackage{graphicx}
\usepackage{booktabs}
\usepackage[table]{xcolor}
\usepackage{tcolorbox}
\usepackage{marvosym}
\usepackage[accsupp]{axessibility}
\usepackage{hyperref}
\usepackage{orcidlink}

\begin{document}

\title{Don't Let the Video Speak:\\ Audio-Contrastive Preference Optimization for Audio-visual Language Models}

\titlerunning{Audio-Contrastive Preference Optimization for AVLMs} 

\author{Ami Baid$^{(\text{\Letter})}$ \and Zihui Xue \and Kristen Grauman}
\authorrunning{Baid et al.}

\institute{University of Texas at Austin, Austin, USA\\
\email{amibaid@utexas.edu}}

\maketitle

\begin{abstract}
    While Audio-Visual Language Models (AVLMs) have achieved remarkable progress over recent years, their reliability is bottlenecked by cross-modal hallucination. A particularly pervasive manifestation is video-driven audio hallucination: models routinely exploit visual shortcuts to hallucinate expected sounds, discarding true auditory evidence. To counteract this deeply ingrained visual dominance, we propose Audio-Contrastive Preference Optimization (ACPO). This dual-axis preference learning framework introduces an output-contrastive objective to penalize visual descriptions masquerading as audio facts, alongside an input-contrastive objective that swaps audio tracks to explicitly penalize generation invariant to the true auditory signal. Extensive experiments demonstrate that ACPO establishes highly faithful audio grounding and mitigates audio hallucination. Project page: \url{https://vision.cs.utexas.edu/projects/acpo/}
\end{abstract}
\section{Introduction}
\label{sec:intro}

\begin{figure}[t]
    \centering
    \includegraphics[width=1\linewidth]{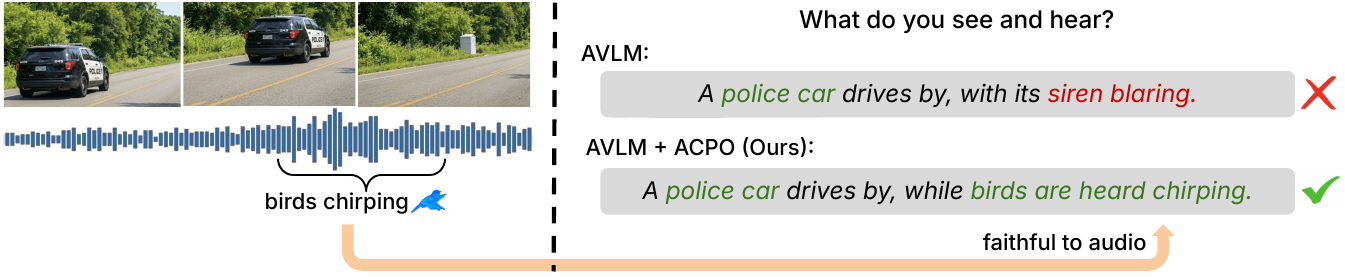}
    \caption{Correcting cross-modal hallucination in AVLMs. Current models often exploit real-world co-occurrence shortcuts, leading them to hallucinate sounds based on what is seen rather than what is heard. Here, the base AVLM incorrectly predicts a siren due to the visual presence of a police car. Our approach explicitly corrects this class of modality attribution errors, decoupling the visual shortcut to accurately ground the response in the actual audio track.}
    \label{fig:concept}
\end{figure}

Large language models (LLMs)~\cite{brown2020language,touvron2023llama,achiam2023gpt} have transformed natural language processing, serving as flexible interfaces for summarization~\cite{stiennon2020learning,lewis2020bart}, question answering~\cite{khashabi2020unifiedqa,lewis2020retrieval}, and reasoning~\cite{wei2022chain,kojima2022large}. But language alone is insufficient for understanding the physical world; human perception seamlessly integrates language, vision, and sound. Vision-language models (VLMs) extend LLMs by incorporating visual input~\cite{alayrac2022flamingo,li2023blip,liu2023visual}. Audio-visual language models (AVLMs) emerged following VLMs, able to jointly reason over text, video, and audio~\cite{videollama2,videosalmonn2,qwen2.5omni}. 

Audio provides critical context that vision alone cannot capture: it alerts us to off-screen events, clarifies the hidden mechanics of object interactions, and conveys the nuanced emotions of speakers. Integrating both auditory and visual streams enables AVLMs to drive meaningful advancements in real-world applications, like autonomous systems~\cite{sima2024drivelm} and assistive technologies~\cite{ainary2025audo,huh2025vid2coach} that must reliably interpret their surroundings. Through generative tasks like question answering and captioning, these models provide a flexible, user-friendly interface to translate complex multimodal environments into accessible insights.

Despite their impressive fluency, LLMs struggle with hallucination, generating text that is plausible but factually ungrounded~\cite{ji2023survey,huang2025survey}. When extended to the visual domain, VLMs inherit this vulnerability and exhibit visual hallucination—confidently describing entities, attributes, or actions that are entirely absent from the visual input even if semantically plausible~\cite{rohrbach2018object,li2023evaluating,guan2024hallusionbench}. Consequently, as models evolve to process increasingly diverse sensory streams, maintaining strict factual grounding across multiple modalities emerges as a central problem.

The integration of audio and video in AVLMs introduces a uniquely challenging variant of this problem: \emph{cross-modal hallucination}. Unlike unimodal errors, these hallucinations are driven by misleading priors across sensory streams, where a model generates claims that are plausible under one modality but unsupported by the full context. Relying on visual priors, for example, a model might hallucinate an accompanying siren for a silent police car (see Fig.~\ref{fig:concept}); conversely, auditory cues might trigger descriptions of off-screen speakers as if they were visible. These errors are deeply insidious; they mimic natural audio-visual co-occurrences and read with perfect fluency. Yet, they expose a critical flaw in the model's modality attribution, undermining the reliability of AVLM generation.

Crucially, cross-modal hallucination is an asymmetric phenomenon. Existing literature demonstrates that AVLMs systematically default to visual priors. Audio tokens receive disproportionately low attention weights during decoding~\cite{avcd}, and models are notably more prone to hallucinating on audio-focused tasks than visual ones~\cite{cmm}. We observe this exact asymmetry in practice (Sec.~\ref{sec:prelim}): feeding more video frames into AVLMs actively triggers more audio hallucinations, while adding audio causes no such degradation to visual QA tasks. This dynamic exposes a deep-seated visual dominance: models are overwhelmingly prone to guessing sounds based on what they see. Addressing this vulnerability is the core focus of our work, as we aim to enforce true auditory grounding and ensure models \emph{don't let the video speak.}

Despite its pronounced impact on model trustworthiness, video-driven audio hallucination is an under-explored vulnerability. Early attempts to address this—spanning both training-free \cite{avcd} and training-based \cite{omnidpo} methods—have proven insufficient, as they either do not penalize cross-modal leakage or leave the weak audio representations unimproved. Deeply investigating this capability gap, we identify two compounding factors at the root of visual dominance. (1) The strong correlation between sight and sound in large-scale datasets \cite{valor, vggsound} encourages models to rely on superficial co-occurrence shortcuts, bypassing the need to learn true modality attribution. (2) The inherent architectural imbalance, where vision encoders benefit from vastly larger datasets and stronger supervision than audio encoders \cite{avhbench, cmm, avcd}, heavily biases the model. As a direct result, visual priors dictate the response, and conflicting audio evidence is systematically ignored~\cite{cmm, avcd}.

To fundamentally correct this visual dominance, we propose Audio-Contrastive Preference Optimization (ACPO). Our key insight is that naively applying preference learning to static audio-visual inputs fails to break co-occurrence shortcuts, as models can still guess the correct audio response using solely visual cues. Instead, ACPO forces true modality attribution by constructing preference pairs along two orthogonal axes. First, we employ output-contrastive pairs using audio-swapped inputs, explicitly penalizing the model for generating visually-driven descriptions in response to audio queries. Second, we introduce input-contrastive pairs that evaluate identical text outputs across different audio tracks, penalizing the model if its predictions remain invariant when the supporting auditory evidence is removed. By jointly optimizing these pairs and fine-tuning only the audio projection layer, our lightweight framework strengthens audio grounding without disrupting the backbone's established vision-language capabilities.

Extensive evaluations validate the efficacy of our framework. To rigorously isolate and measure modality-specific grounding in free-form generation, we introduce a novel unimodal captioning evaluation that probes audio and visual understanding under both aligned and mismatched conditions. Across this rigorous new protocol and established standard hallucination benchmarks, ACPO elevates isolated audio captioning quality and decisively curtails video-driven audio hallucinations, while preserving general multimodal capabilities. Moreover, these gains hold across three distinct AVLM architectures, indicating ACPO addresses a general failure mode of audio-visual grounding. Ultimately, our findings establish that targeted, contrastive modality supervision is a highly effective and necessary mechanism for achieving balanced, trustworthy response generation in AVLMs.
\section{Related Work}

\subsection{Audio-visual Language Models}
AVLMs extend vision-language models (VLMs) by incorporating audio as an additional input stream \cite{videollama, panda, videosalmonn2}. In a typical architecture, a video encoder \cite{clip, siglip} and an audio encoder \cite{beats} produce modality-specific embeddings, which are projected into a shared space and fed alongside tokenized text into an LLM backbone. Early AVLMs \cite{videollama, panda} demonstrated the viability of this paradigm, and more recent models \cite{videollama2, videosalmonn2, qwen2.5omni} achieve competitive performance on benchmarks requiring joint audio-visual reasoning \cite{valor, avqa, musicavqa}.

\subsection{Hallucination in LLMs and VLMs}
Hallucination has been studied extensively in LLMs~\cite{ji2023survey,huang2025survey} and VLMs~\cite{rohrbach2018object,li2023evaluating,guan2024hallusionbench,thinking-drift, arrow-of-time, progress-aware}. Training free strategies include chain-of-thought prompting~\cite{wei2022chain}, self consistency sampling\cite{mrfd}, and contrastive decoding methods\cite{vcd}, which adjust outputs' logits to reduce reliance on language priors. Training-based approaches include hallucination-aware supervised objectives \cite{dftg, reverse}, auxiliary alignment losses \cite{mma}, and preference based learning \cite{dpo, vdpo}. DPO \cite{dpo} has emerged as a particularly effective framework, able to steer model outputs without catastrophic forgetting \cite{beyond}. Unlike policy-gradient methods such as PPO\cite{ppo}, DPO operates on fixed preference pairs without requiring an explicit reward model or on-policy rollouts. V-DPO\cite{vdpo} applies DPO to visual hallucination by constructing preference pairs using out-of-distribution objects (e.g., cutting a rock instead of a cake). These methods are effective for vision-language grounding, but they address only a single non-text modality and do not consider interactions between multiple input streams.

\subsection{Cross-Modal Hallucination in AVLMs}
Cross-modal hallucination in AVLMs remains relatively underexplored. Early work~\cite{nishimura} offers analysis of audio hallucination in AVLMs, finding that models frequently generate audio descriptions driven by visual content rather than the audio signal. Subsequent benchmarks probe this failure more systematically. AVHBench \cite{avhbench} evaluates cross-modal hallucination using questions whose ground truth response depends on a single modality while the other may be misleading. CMM \cite{cmm} decomposes hallucination into unimodal dominance and spurious inter-modality correlations, measuring both under controlled masking and corruption. Both benchmarks confirm that current AVLMs are especially vulnerable when audio and video conflict.

To address this, audio-visual contrastive decoding \cite{avcd} masks selected modalities at inference time and adjusts logits to reduce hallucination, but incurs additional cost and does not modify underlying representations. OmniDPO \cite{omnidpo} extends preference optimization to the audio-visual setting. For each example, they add noise to the audio or video stream and train the model to prefer responses conditioned on the clean input over the corrupted one. However, noisy inputs remain broadly aligned with the original video, and therefore do not directly address video-driven audio hallucination. Our approach differs by constructing preference pairs with cross-modal conflict that elicit hallucination, and by introducing modality-specific supervision that penalizes visually hallucinated audio descriptions.

\section{Method}

\begin{figure}[t]
    \centering
    \includegraphics[width=1\linewidth]{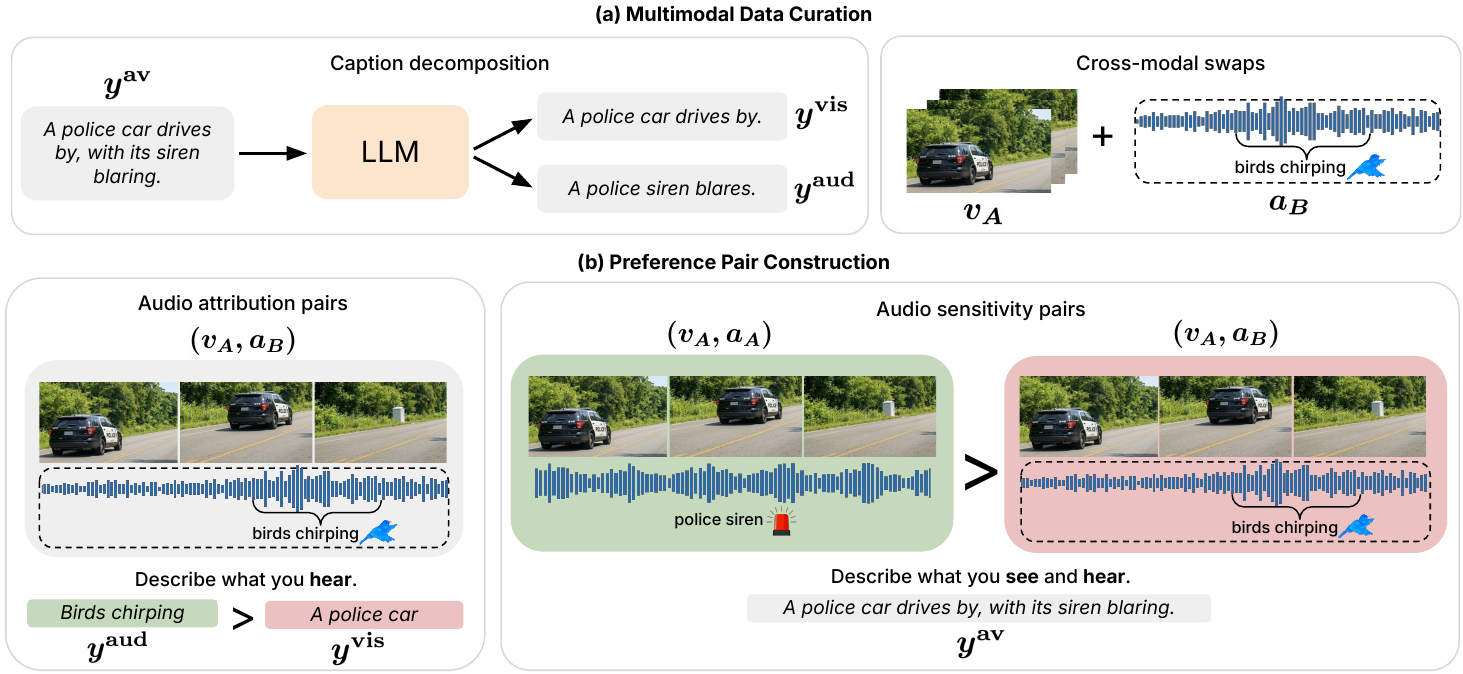}
    \caption{Overview of ACPO. (a) Multimodal Data curation: each joint audio-visual caption is decomposed into modality-specific targets $y^\text{vis}$ and $y^\text{aud}$, and audio-swapped inputs $(v_A, a_B)$ are constructed by replacing the original audio track with a mismatched one. (b) Preference pair construction: audio-attribution pairs (left) use the swapped input $(v_A, a_B)$ to penalize visually-driven responses to audio-focused prompts, preferring $y^\text{aud}$ over $y^\text{vis}$. Audio-sensitivity pairs (right) penalize audio-invariant predictions by preferring the original audio-visual caption $y^\text{av}$ under the aligned input $(v_A, a_A)$ over the same caption under the swapped input $(v_A, a_B)$.}
    \label{fig:method}
\end{figure}

We introduce Audio-Contrastive Preference Optimization (ACPO), a lightweight preference-based framework that strengthens audio grounding in AVLMs. We describe the setup and background in Sec. 3.1, and detail ACPO in Sec. 3.2.

\subsection{Problem Setup}
\label{sec:prelim}

We consider an AVLM that receives a video $v$, an audio track $a$, and a text prompt $x$, and generates a text response $y$. The model $p_\theta(y \mid v, a, x)$ encodes $v$ and $a$ through modality-specific encoders, projects the resulting representations into a shared embedding space via learned projection modules, and feeds the projected tokens alongside the tokenized prompt into an LLM backbone. 
Our objective is to ensure that the model intelligently integrates cues from both $v$ and $a$, maintaining factual grounding across both modalities rather than hallucinating content from one single modality. 

\begin{figure}[t]
    \centering
    \includegraphics[width=1\linewidth]{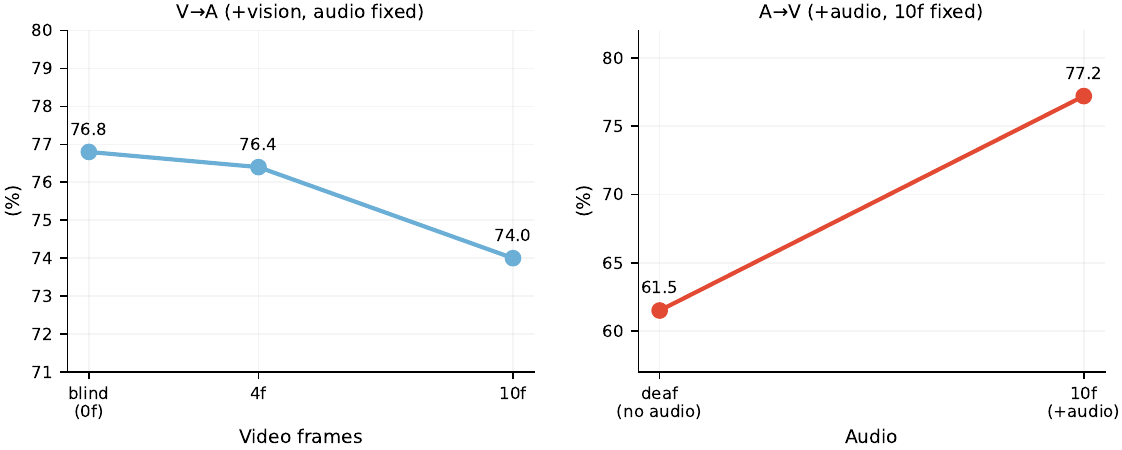}
    \caption{Visual dominance in AVLMs is asymmetric. On the AVHBench \cite{avhbench} video-driven audio hallucination task (V$\rightarrow$A), adding more video frames progressively degrades performance (76.8 $\rightarrow$ 74.0), as visual priors override auditory evidence. On the audio-driven video hallucination task (A$\rightarrow$V), adding audio improves performance (61.5 $\rightarrow$ 77.2), indicating the model is robust to audio interference but highly susceptible to visual dominance.}
    \label{fig:av_analysis}
\end{figure}

Consistent with prior findings~\cite{avcd, avhbench, cmm}, we observe an asymmetry in modality reliance, where models disproportionately favor visual cues. This imbalance reflects a capability gap, as vision encoders and visual training pipelines are typically stronger and more mature than their audio counterparts. As shown in our analysis (Fig.~\ref{fig:av_analysis}, left), when evaluating audio-focused question answering, progressively adding visual information (by scaling the number of input frames) actively degrades performance. This suggests that rather than providing complementary context, the added visual input confuses the model and overrides valid audio evidence. Conversely, augmenting video-focused QA tasks with the audio modality does not result in a similar performance drop (right), indicating that the model is relatively robust to audio interference but highly susceptible to visual dominance.

Motivated by this vulnerability to visual interference, we introduce Audio-Contrastive Preference Optimization (ACPO). ACPO is designed to counteract visual dominance, ensuring that the model remains faithfully grounded in the actual audio content instead of hallucinating sounds driven by strong visual priors.

\subsection{Audio-Contrastive Preference Optimization}
\label{sec:acpo}

Correcting visual dominance requires teaching the model to trust the audio signal even when the visual signal provides a tempting, but potentially misleading, prior. We formulate this as a preference learning problem. Our foundation is Direct Preference Optimization (DPO)~\cite{dpo}, which aligns language models by contrasting a preferred response $y^+$ against a dispreferred response $y^-$ conditioned on an input context $c$:
\begin{equation}
\label{eq:dpo_standard}
\mathcal{L}_{\text{DPO}} = -\log\sigma\!\Big(\beta\big[\log\tfrac{p_\theta(y^+\mid c)}{p_{\mathrm{ref}}(y^+\mid c)} - \log\tfrac{p_\theta(y^-\mid c)}{p_{\mathrm{ref}}(y^-\mid c)}\big]\Big),
\end{equation}
where $p_{\mathrm{ref}}$ is the frozen reference model and $\beta$ controls the deviation penalty. Intuitively, DPO increases the relative likelihood of the preferred response while anchoring updates to the reference distribution.

Our key insight is that naively applying standard DPO—which relies on a fixed, joint audio-visual context $c$—fails to isolate the root cause of video-driven hallucination. Because real-world audio and video are highly correlated, a model can learn to output the ``preferred'' audio caption merely by exploiting visual co-occurrence shortcuts, defeating the purpose of the alignment. To force true audio grounding, ACPO instead constructs preference pairs along two axes: one contrasts \emph{what the model says} (output-contrastive), the other contrasts \emph{what the model hears} (input-contrastive). The pair types rely on modality-specific supervision targets and controlled audio--visual mismatches, which we describe first. Fig.~\ref{fig:method} illustrates both the data curation pipeline and the resulting preference pairs.

\textbf{Caption decomposition.}
Standard audio-visual captions entangle visual and auditory evidence in a single sentence, offering no signal for modality-specific grounding. Starting from standard audio-visual captions, we use a large language model to decompose each joint caption into a visual caption $y^{\mathrm{vis}}$ describing only visually verifiable content and an audio caption $y^{\mathrm{aud}}$ describing only audibly verifiable content. These decomposed captions serve as modality-specific supervision targets for the preference pairs below. 

\textbf{Audio-swapped inputs.}
To decouple audio from its co-occurring visual context, we construct audio-swapped inputs that retain the original video while replacing the audio track. For a source clip with video $v_A$ and audio $a_A$, we sample a partner clip and form the swapped input $(v_A, a_B)$. To control the degree of audio-visual mismatch, we measure the similarity between the original video and candidate audio tracks using a multimodal embedding model \cite{imagebind}, grouping swaps into high- and low-similarity tiers based on empirical quantiles, where low-similarity swaps introduce stronger audio-visual conflict.

\textbf{Audio-attribution pairs (output-contrastive).}
These pairs penalize the model for describing visual content in response to an audio-specific prompt. For a clip with video $v_A$ and swapped track $a_B$, we prompt the model with an audio-focused instruction $x_{\mathrm{aud}}$ (e.g., ``Describe what you hear.''). The preferred response is the audio caption $y_B^{\mathrm{aud}}$ corresponding to the actual audio track $a_B$; the dispreferred response is the visual caption $y_A^{\mathrm{vis}}$ for the video $v_A$. The visual caption may be fluent and plausible given the video, but it reflects visual rather than auditory evidence. These pairs directly penalize the model for sourcing its response from the wrong modality, enforcing the preference:
\begin{equation}
\log p_\theta(y_B^{\mathrm{aud}} \mid v_A, a_B, x_{\mathrm{aud}}) > \log p_\theta(y_A^{\mathrm{vis}} \mid v_A, a_B, x_{\mathrm{aud}}).
\end{equation}

\textbf{Audio-sensitivity pairs (input-contrastive).}
These pairs penalize the model when its predictions are invariant across different audio tracks, indicating that it is not using auditory evidence. For a fixed video $v_A$, we consider both the aligned input $(v_A, a_A)$ and an audio-swapped input $(v_A, a_B)$. Let $y_A^{\mathrm{av}}$ denote the original audio-visual caption for clip $A$, and let $x$ be a joint instruction (e.g., ``Describe what you see and hear.''). The preferred configuration is $((v_A, a_A),\; y_A^{\mathrm{av}})$; the dispreferred configuration is $((v_A, a_B),\; y_A^{\mathrm{av}})$. Since $a_B$ does not support the audio-specific content in $y_A^{\mathrm{av}}$, the model should assign lower likelihood to the same caption when the supporting auditory evidence has been replaced. These pairs penalize audio-insensitive behavior, enforcing the preference:
\begin{equation}
\log p_\theta(y_A^{\mathrm{av}} \mid v_A, a_A, x) > \log p_\theta(y_A^{\mathrm{av}} \mid v_A, a_B, x).
\end{equation}

 During training, we optimize the DPO objective (Eq.~\ref{eq:dpo_standard}) over the union of all pair types with lightweight fine-tuning to improve audio grounding.
\section{Experiments}

\begin{figure} [t]
    \centering
    \includegraphics[width=1\linewidth]{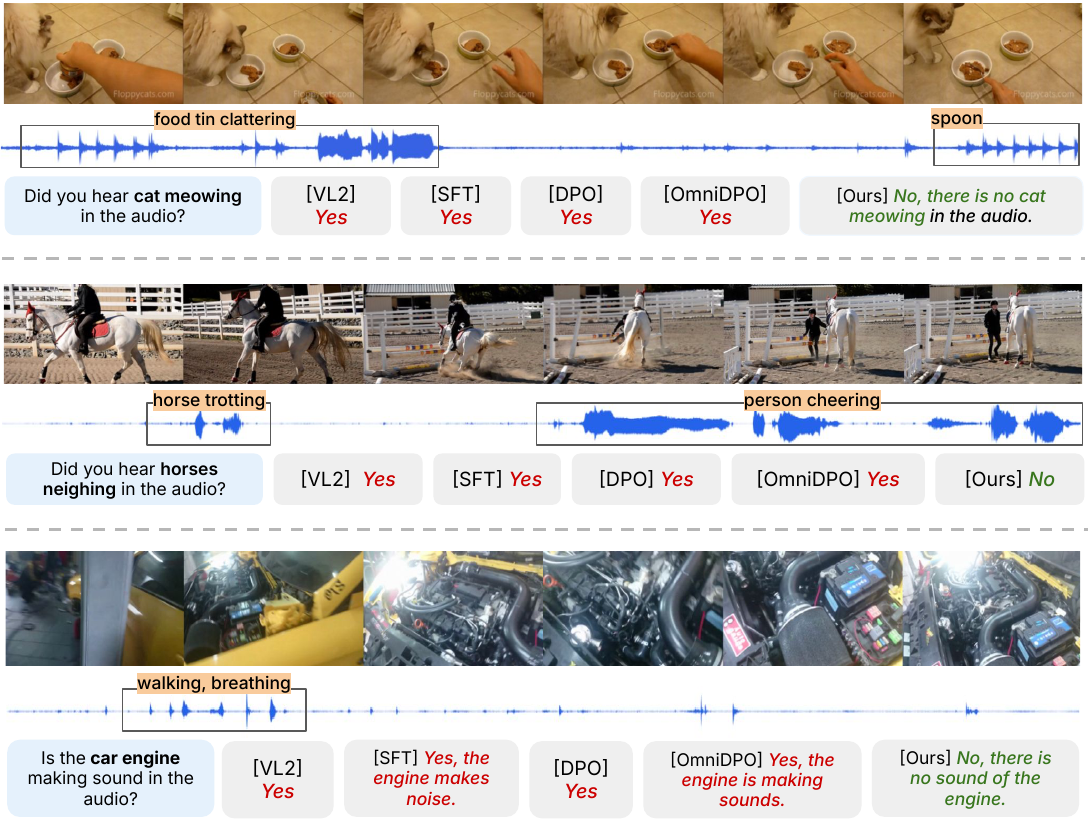}
    \caption{Qualitative examples of video-driven audio hallucination. Each row shows a video clip (top), its corresponding audio waveform with labeled sound events (middle), and model responses to an audio-focused yes/no question (bottom). In all three cases, the audio contains no evidence of the queried sound, yet all baselines hallucinate affirmative responses. ACPO (Ours) correctly grounds its response in the audio signal.}
     \label{fig:qual1}
\end{figure}

\begin{figure}[t]
    \centering
    \includegraphics[width=1\linewidth]{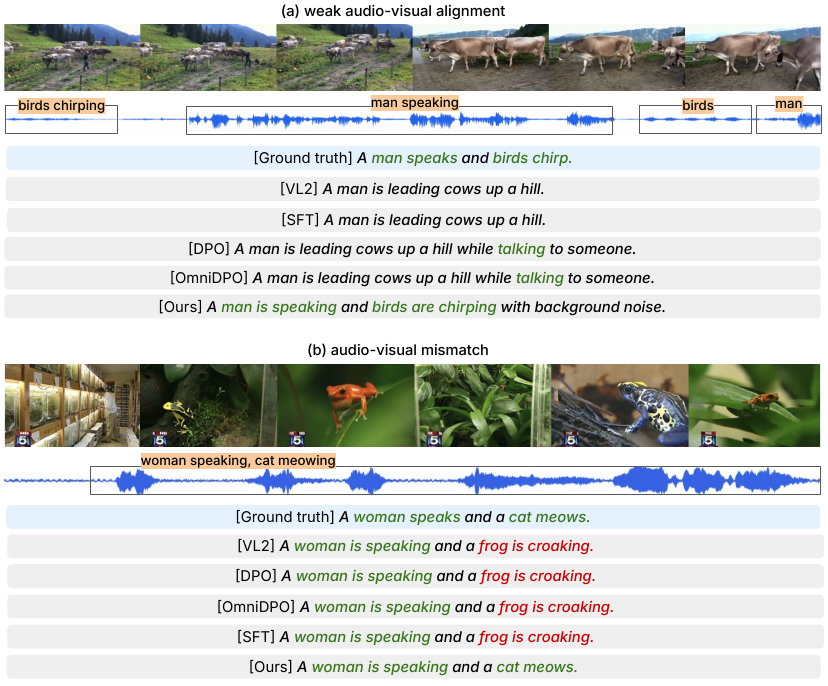}
    \caption{Qualitative examples of audio-focused captioning. Each row shows a video clip with its labeled audio waveform, a reference audio caption, and model-generated audio captions. (a) The video depicts cows on a hillside, and the audio contains a man speaking and birds chirping. All baselines produce captions grounded in the visual scene (``leading cows up a hill''), failing to describe the actual audio content. ACPO correctly identifies both auditory events. (b) The video shows frogs, but the audio contains a woman speaking and a cat meowing. All baselines hallucinate a frog croaking. ACPO alone correctly describes what is heard.}
     \label{fig:qual2}
\end{figure}

We describe the benchmarks and datasets we use to validate our approach, overview implementation details and baselines, and present our results.

\subsection{Setup}

\textbf{Benchmarks.}
AVHBench~\cite{avhbench} specifically probes cross-modal hallucination in AVLMs. The benchmark consists of yes/no questions whose ground truth depends on a single modality, while the other may be misleading. We report results on the video-driven audio (V$\rightarrow$A) hallucination task, the audio-driven video hallucination (A$\rightarrow$V) task, and audio-visual captioning. CMM~\cite{cmm} evaluates unimodal dominance and multimodal robustness under controlled modality masking and corruption. We evaluate on four tasks: Audio-Language, Overreliance on Vision, Overreliance on Audio, and Vision-Audio-Language. We do not include vision-language tasks, since we only modify the audio projector.

\textbf{Unimodal Captioning Evaluation.} Open-ended video captioning is highly susceptible to cross-modal hallucination, yet evaluating modality-specific grounding remains a challenge. Standard audio-visual datasets typically provide joint captions that are largely vision-focused. Furthermore, datasets that do offer unimodal annotations \cite{vast} often contain generic and uninformative audio descriptions (e.g., ``someone is speaking''). This makes it difficult to isolate a model's true audio understanding from its reliance on visual shortcuts. To help close this gap, we constructed a targeted evaluation set to explicitly measure decoupled audio-visual understanding. We sourced videos and their corresponding joint captions from the AVHBench audio-visual captioning task, filtering for 400 clips that contain diverse and distinct audio events. To evaluate robustness against hallucination, we also constructed 400 audio-swapped counterparts. We used an LLM to rank candidate audio tracks to select plausible mismatches. To generate ground truth audio captions, we provided an AVLM \cite{gemini25} with the original multimodal caption alongside the raw audio track, instructing it to describe only the verifiable auditory events. Conversely, to generate the vision-only ground truth, we prompted the model with the original caption and the raw video frames. We manually verified a random 10\% of the generated captions, of which 95\% were accurate in both factual content and modality attribution. Note that during evaluation, all models receive both video and audio as input. This mirrors the cross-modal hallucination setting: the model must describe what it hears despite the presence of potentially misleading visual input. The unimodal ground-truth captions serve as modality-specific reference targets for measuring grounding.

\textbf{Training Data.} We construct training pairs from 5,000 VALOR~\cite{valor} clips with joint captions. We decompose each caption into modality-specific targets using GPT-5\cite{gpt5} and construct audio-swapped pairs as described in Sec.~\ref{sec:acpo}. We combine our audio-contrastive pairs with the noise-based and text DPO pairs from~\cite{omnidpo}, sampling randomly within each batch at a 60/40 ratio of audio-contrastive to other multimodal pairs.

\textbf{Implementation Details.} We explore ACPO's impact across three AVLM backbones. We adopt Video-LLaMA2-7B-AV~\cite{videollama2} as  our primary backbone, given its leading performance on the AVHBench and CMM leaderboards. We additionally apply ACPO to video-SALMONN-2~\cite{videosalmonn2} and Qwen2.5-Omni~\cite{qwen2.5omni}~(Sec.~4.3). The training configuration below is shared across all three backbones, differing only in the parameter-efficient fine-tuning scope. For Video-LLaMA2, we freeze the video encoder, audio encoder, and the LLM backbone. We fine-tune only the audio projection layer to strengthen audio-language alignment without disrupting the model's established vision-language capabilities. We train for one epoch over 5,000 preference pairs using AdamW with a learning rate of 2e-5 and cosine scheduling with warmup, a DPO $\beta$ of 0.1, and a global batch size of 8. Training beyond one epoch yields no additional gain. Training completes in approximately 3 hours on a single NVIDIA GH200 120GB GPU.

\textbf{Baselines.} We compare ACPO against the following established training strategies. All baselines are implemented on the same Video-LLaMA2 backbone and follow the same training paradigm as ours, ensuring a fair comparison.

\begin{enumerate}
    \item Base AVLM (zero-shot): Unmodified pretrained Video-LLaMA2-7B-AV~\cite{videollama2}.
    \item SFT~\cite{ouyang2022training}: Fine-tuned on VALOR captions using standard cross-entropy loss.
    \item DPO~\cite{dpo}: Preference pairs where audio-visual responses are preferred over vision-only responses on original VALOR clips.
    \item \textbf{OmniDPO}~\cite{omnidpo}: The closest prior preference optimization method for omni-modal hallucination, using text preference pairs and noise-based input corruption. We reimplement it within our training setup.
\end{enumerate}

Additionally, to provide broader context for our results, we include zero-shot evaluations of several state-of-the-art AVLMs:
Gemini-Flash-1.5~\cite{gemini15}, Qwen2.5-Omni~\cite{qwen2.5omni}, and MiniCPM-o-2.6~\cite{minicpm}. These models are not directly comparable to our controlled baselines, but provide 
reference points for situating our results.

\subsection{Main Results}

\begin{table} [t]
\centering
\caption{Correcting audio hallucination on two public benchmarks. Our method achieves the best overall performance on both benchmarks, demonstrating superior audio hallucination resistance and discrimination. PA: accuracy on yes-instances, HR: accuracy on no-instances. Best in bold. $^*$Results reported in cited work.}
\label{tab:audio}
\scriptsize
\setlength{\tabcolsep}{3pt}
\resizebox{\columnwidth}{!}{%
\begin{tabular}{lcccccccccccccc}
\toprule
& \multicolumn{6}{c}{\textbf{AVHBench}} & \multicolumn{7}{c}{\textbf{CMM}} \\
\cmidrule(lr){2-7}\cmidrule(lr){8-14}
& \multicolumn{4}{c}{Audio Hallucination} & \multicolumn{2}{c}{Overall} & \multicolumn{3}{c}{Aud-Lang} & \multicolumn{3}{c}{Overrely Vision} & \multicolumn{1}{c}{Overall} \\
\cmidrule(lr){2-5}\cmidrule(lr){6-7}\cmidrule(lr){8-10}\cmidrule(lr){11-13}\cmidrule(lr){14-14}
\textbf{Method} & Prec$\uparrow$ & Rec$\uparrow$ & F1$\uparrow$ & Acc$\uparrow$ & F1$\uparrow$ & Acc$\uparrow$ & PA$\uparrow$ & HR$\uparrow$ & Acc$\uparrow$ & PA$\uparrow$ & HR$\uparrow$ & Acc$\uparrow$ & Acc$\uparrow$ \\
\midrule
\rowcolor{gray!20}Gemini-Flash-1.5$^*$~\cite{gemini15} & 57.9 & 94.7 & 71.9 & 63.0 & 77.8 & 73.2 & 88.5 & 39.5 & 64.0 & 79.0 & 36.5 & 57.8 & 76.3 \\
\rowcolor{gray!20}Qwen2.5-Omni$^*$~\cite{omnidpo} & 60.8 & 98.8 & 75.3 & 67.6 & 76.4 & 70.9 & 92.0 & 78.0 & 85.0 & 95.0 & 56.5 & 75.8 & 81.0 \\
\rowcolor{gray!20}MiniCPM-o$^*$~\cite{omnidpo} & 70.4 & 78.6 & 74.4 & 72.8 & 75.1 & 73.7 & 95.0 & 53.0 & 74.0 & 91.0 & 56.5 & 73.8 & 76.0 \\
\midrule
Base model \cite{videollama2} & 68.6 & 88.5 & 77.3 & 74.0 & 77.6 & 75.6 & 85.5 & 85.5 & \textbf{85.5} & 82.0 & 64.5 & 73.3 & 82.5 \\
SFT~\cite{ouyang2022training} & 72.8 & 83.5 & 77.8 & 76.2 & 78.1 & 77.0 & 84.0 & 86.0 & 85.0 & 80.5 & 71.5 & 76.0 & 82.5 \\
DPO \cite{dpo} & 70.6 & \textbf{88.9} & 78.7 & 76.0 & 78.5 & 76.7 & \textbf{87.0} & 82.5 & 84.8 & \textbf{82.5} & 72.0 & 77.3 & 82.9 \\
OmniDPO~\cite{omnidpo} & 76.6 & 81.6 & 79.0 & 78.4 & 78.7 & 78.6 & 84.5 & 81.5 & 83.0 & 77.0 & 77.5 & 77.3 & 82.4 \\
\rowcolor{green!18}Ours & \textbf{78.2} & 82.8 & \textbf{80.4} & \textbf{79.9} & \textbf{79.2} & \textbf{78.8} & 84.5 & \textbf{86.5} & \textbf{85.5} & 80.5 & \textbf{82.0} & \textbf{81.3} & \textbf{83.4} \\
\bottomrule
\end{tabular}}%
\end{table}
\begin{table} [t]
\centering
\caption{Unimodal captioning evaluation on our newly introduced eval set. Our method achieves the largest gains on audio captioning while remaining competitive on video. Each cell shows METEOR/CIDEr (×100), $\uparrow$. Because each split uses modality-specific references, these metrics capture fidelity to the intended modality. Original: matched audio-video pairs; Swap: mismatched but plausible substitutions. Best in bold.}
\label{tab:captioning}
\small
\setlength{\tabcolsep}{5pt}
\begin{tabular}{lccccc}
\toprule
& Audio & Audio & Video & Video & Average \\
\textbf{Method} & (Original) & (Swap) & (Original) & (Swap) &  \\
\midrule
Base model~\cite{videollama2} & 18.8/27.9 & 15.0/13.1 & 22.7/28.5 & \textbf{23.2}/29.7 & 19.9/24.8 \\
SFT~\cite{ouyang2022training} & 22.4/31.3 & 17.3/16.9 & 19.9/\textbf{33.6} & 20.9/\textbf{40.4} & 20.1/30.5 \\
DPO~\cite{dpo} & 22.4/27.6 & 18.0/16.5 & \textbf{22.9}/23.3 & 23.2/27.1 & 21.6/23.6 \\
OmniDPO~\cite{omnidpo} & 23.1/33.0 & 18.9/18.9 & 21.9/25.1 & 21.4/26.1 & 21.3/25.8 \\
\rowcolor{green!18}Ours & \textbf{27.0}/\textbf{43.6} & \textbf{23.3}/\textbf{30.6} & 20.4/27.6 & 20.2/29.7 & \textbf{22.7}/\textbf{32.9} \\
\bottomrule
\end{tabular}
\end{table}
\begin{table} [t]
\centering
\caption{ACPO transfers to architectures beyond Video-LLaMA2. Applied to video-SALMONN-2 and Qwen2.5-Omni, ACPO reduces audio hallucination and improves overall AVHBench performance on both, suggesting that its dual-axis preference signal targets a general grounding failure rather than a single-backbone artifact. Best per backbone in bold.}
\label{tab:generalization}
\setlength{\tabcolsep}{8pt}
\begin{tabular}{lcccc}
\toprule
 & \multicolumn{2}{c}{Audio Halluc.} & \multicolumn{2}{c}{Overall} \\
\cmidrule(lr){2-3}\cmidrule(lr){4-5}
Model & Acc$\uparrow$ & F1$\uparrow$ & Acc$\uparrow$ & F1$\uparrow$ \\
\midrule
video-SALMONN-2 (base)~\cite{videosalmonn2} & 63.2 & 54.0 & 71.6 & 67.0 \\
\quad + ACPO & \textbf{68.8} & \textbf{71.6} & \textbf{75.0} & \textbf{76.5} \\
\midrule
Qwen2.5-Omni (base)~\cite{qwen2.5omni} & 66.7 & 74.8 & 68.4 & 74.8 \\
\quad + ACPO & \textbf{69.3} & \textbf{76.2} & \textbf{71.0} & \textbf{76.1} \\
\bottomrule
\end{tabular}
\end{table}

\textbf{Audio hallucination.} ACPO is the strongest method at resisting audio hallucination across both benchmarks (Table~\ref{tab:audio}). On AVHBench audio hallucination, ACPO obtains the highest F1 (80.4) and accuracy (79.9). The baseline exhibits high recall but low precision, indicating a tendency to over-predict plausible sounds. ACPO shifts toward higher precision while maintaining competitive recall. On CMM, ACPO achieves 82.0 hallucination resistance on Overrely Vis, a substantial improvement over the baseline (64.5) and the next best method, OmniDPO (77.5). This subcategory measures robustness when visual content could mislead audio predictions, precisely the failure mode ACPO targets, as Fig.~\ref{fig:qual1} illustrates. ACPO also achieves the highest overall accuracy on both Overrely Vis (81.3) and Aud-Lang (85.5). Aud-Lang evaluates audio understanding without visual input, and ACPO is the only method that does not degrade performance on this subcategory relative to the baseline.

\textbf{Captioning.}
ACPO produces the largest gains on audio captioning (Table~\ref{tab:captioning}). Audio CIDEr improves from 27.9 to 43.6 on original clips and from 13.1 to 30.6 on swapped clips, substantially outperforming all baselines. These gains demonstrate that ACPO improves the model's ability to describe audio content, not just its ability to reject hallucinated descriptions. The swapped-clip results are particularly informative: when audio and video conflict, ACPO correctly describes what it hears rather than defaulting to visual content, as Fig.~\ref{fig:qual2} illustrates. Video captioning scores also remain competitive with DPO and OmniDPO, decreasing only slightly relative to the baseline, which is expected as the baseline is heavily visually dominant. Overall, ACPO produces a more balanced model that achieves the best average across all conditions on both METEOR and CIDEr.

\textbf{Multimodal performance.}
ACPO improves audio grounding with minimal impact on broader multimodal capabilities, achieving the best overall accuracy on both AVHBench (78.8) and CMM (83.4). Table~\ref{tab:audio} reports audio-focused performance; full results on remaining tasks are provided in the supplementary material. Scores on video-focused tasks shift only slightly relative to the base model. These shifts reflect rebalancing rather than degradation. The base model is visually dominant, and its high video-captioning scores partly reflect a tendency to produce purely visual captions that trivially match video-only references. ACPO trades a small amount of this visual bias for substantially improved audio grounding, gains that far outweigh the modest video-side cost. Joint audio-visual capability is preserved on AVHBench joint captioning (supplementary material).

\subsection{Generalization Across AVLM Backbones}

ACPO's gains are not specific to Video-LLaMA2~\cite{videollama2}. Applied to two further open-source AVLMs, video-SALMONN-2~\cite{videosalmonn2} and Qwen2.5-Omni~\cite{qwen2.5omni}, it improves both the targeted audio-hallucination task and overall performance on each (Table~\ref{tab:generalization}).
We keep the ACPO objective and the same 5{,}000 preference pairs unchanged and train for one epoch as in our main setup, adapting only the parameter-efficient fine-tuning scope to each architecture. For these two backbones we additionally apply low-rank adaptation (LoRA) to the LLM alongside the audio projection module. These consistent gains across three distinct AVLMs suggest that ACPO addresses a general failure mode of audio-visual grounding rather than a single-backbone artifact.

\subsection{Ablations}

\begin{table} [t]
\centering
\caption{Pair type ablation on the AVHBench audio hallucination task.}
\label{tab:ablation-pair}
\setlength{\tabcolsep}{3pt}
\begin{tabular}{l cccc}
\toprule
\textbf{Configuration} & Acc$\uparrow$ & Prec$\uparrow$ & Rec$\uparrow$ & F1$\uparrow$ \\
\midrule
Ours  & \textbf{79.9} & 78.2 & 82.8 & \textbf{80.4} \\
w/o Attribution     & 79.1 & 76.2 & \textbf{84.6} & 80.2 \\
w/o Sensitivity     & 79.3 & \textbf{78.3} & 81.0 & 79.6 \\
\bottomrule
\end{tabular}
\end{table}
Table~\ref{tab:ablation-pair} validates the contribution of each pair type, with our full model achieving the best accuracy and F1, indicating the best overall balance between precision and recall. Audio-attribution pairs improve precision, reducing the model's tendency to confirm hallucinated sounds, while audio-sensitivity pairs improve recall, reflecting better attention to the actual audio signal. Together they yield the best overall F1.

\begin{table} [t]
\centering
\caption{Similarity score ablation on the AVHBench audio hallucination task. Each row trains with a single pair type in isolation. Low-similarity swaps introduce stronger audio-visual conflict; high-similarity swaps produce subtler mismatches. $^\dagger$Configurations where video hallucination task accuracy fell below the base model, indicating overcorrection toward audio.}
\label{tab:sim-ablation}
\begin{tabular}{l cccc}
\toprule
\textbf{Pair Type} & Acc$\uparrow$ & Prec$\uparrow$ & Rec$\uparrow$ & F1$\uparrow$ \\
\midrule
Audio-attribution (low sim swap)  & \textbf{77.7} & 76.0 & \textbf{81.1} & \textbf{78.4} \\
Audio-attribution (high sim swap) & 77.0 & 74.9 & \textbf{81.1} & 77.9 \\
Audio-attribution (no swap)       & 76.9 & \textbf{77.6} & 75.7 & 76.6 \\
\midrule
Audio-sensitivity (low sim swap)$^\dagger$  & 79.4 & \textbf{78.9} & 80.1 & 79.5 \\
Audio-sensitivity (high sim swap)$^\dagger$ & \textbf{79.8} & \textbf{78.9} & \textbf{81.3} & \textbf{80.1} \\
\bottomrule
\end{tabular}
\end{table}
Table~\ref{tab:sim-ablation} ablates the effect of audio-visual similarity in the swapped pairs, training each pair type in isolation. For audio-attribution pairs, low-similarity swaps yield the best accuracy and F1. Notably, the no-swap variant achieves the lowest accuracy and F1, underscoring that audio-swapped inputs are essential to the effectiveness of attribution pairs. Without them, the audio and video remain aligned, allowing the model to arrive at the preferred audio caption through visual shortcuts alone and weakening the training signal. For audio-sensitivity pairs, the pattern reverses: high-similarity swaps perform best, likely because subtler mismatches produce harder negatives that better expose audio-invariant behavior. However, both sensitivity-only configurations cause a drop in video hallucination accuracy, suggesting that input-contrastive training in isolation can overcorrect toward audio reliance. The full ACPO objective avoids this trade-off.

\begin{figure} [t]
    \centering
    \includegraphics[width=1\linewidth]{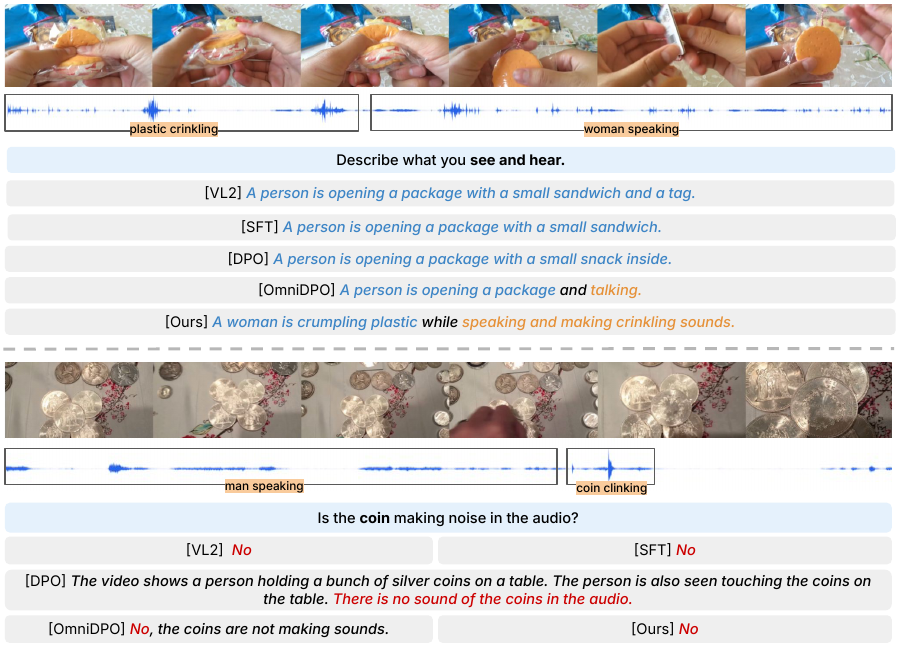}
    \caption{Qualitative examples illustrating limitations. (Top) Joint audio-visual captioning: ACPO provides the most complete multimodal description but generalizes visual content in favor of auditory detail. (Bottom) Audio-focused question answering: all methods fail to detect a brief, subtle coin clinking sound.}
    \label{fig:limit}
\end{figure}

\subsection{Limitations}

Figure~\ref{fig:limit} illustrates two remaining limitations. In the first example, models are asked to describe what they see and hear. The base model, SFT, and DPO produce purely visual descriptions, ignoring the audio entirely. OmniDPO partially captures audio but drops visual details. ACPO correctly identifies both speech and crinkling sounds. However, in this case, ACPO prioritizes auditory cues, generalizing visual content in the process (e.g., package with a sandwich $\rightarrow$ crumpling plastic). In the second example, the audio contains a brief coin clinking sound alongside speech. All methods, including ACPO, fail to detect it. This suggests that brief, subtle audio events remain a challenge even with improved audio grounding.

\section{Conclusion}
In this work, we address the critical challenge of cross-modal hallucination in AVLMs, a pervasive issue where strong visual priors override auditory evidence and cause models to confidently fabricate sounds. To counteract this visual dominance, we propose ACPO, a tailored preference learning framework, which forces the model to decouple spurious audio-visual shortcuts and learn accurate modality attribution. Extensive experiments demonstrate that our approach greatly mitigates video-driven audio hallucination with minimal impact on the model's overarching multimodal capabilities. Ultimately, this work provides a crucial step toward achieving balanced and trustworthy audio-visual understanding.

\section*{Acknowledgements}

This research is supported in part by the UT Austin IFML NSF AI Institute and Lyda Hill Philanthropies. We thank the UT Austin Vision Group for helpful discussions.

\bibliographystyle{splncs04}
\bibliography{main}

\clearpage
\appendix
\begin{center}
{\Large\bfseries Supplementary Material}
\end{center}

\section{Prompt Templates}
We provide the prompts used for caption decomposition and evaluation. Variables are shown in \texttt{monospace}.

\subsection{Training Data: Caption Decomposition}

\begin{tcolorbox}[title=VALOR Caption Splitting (GPT-5), colback=gray!5, colframe=gray!50]
\small
You split captions into two disjoint captions for a video.
\begin{itemize}
    \item video\_caption: ONLY facts that are explicitly visual. Keep it direct and factual.
    \item audio\_caption: ONLY facts that are explicitly audible. Remove visual details (like colors).
\end{itemize}
You will receive a JSON array of items as input. For each item, produce an output JSON object:\\
\texttt{\{"id":"<same id>", "video\_caption":"...", "audio\_caption":"..."\}}\\
\end{tcolorbox}

\subsection{Evaluation: Unimodal Caption Generation}

\begin{tcolorbox}[title=Audio Caption Generation (Gemini 2.5 Pro), colback=gray!5, colframe=gray!50]
\small
You are given an audio clip extracted from a video, along with a reference description: \texttt{\{caption\}}.\\[4pt]
Write a single concise sentence describing the sound events in the audio. Name the sounds (e.g., ``a dog barks'', ``bees buzz'', ``a man speaks'') but do NOT transcribe speech or describe sounds in fine detail. Keep it brief.
\end{tcolorbox}

\begin{tcolorbox}[title=Video Caption Generation (Gemini 2.5 Pro), colback=gray!5, colframe=gray!50]
\small
You are given a muted video clip (audio removed), along with a reference description: \texttt{\{caption\}}.\\[4pt]
Write a single concise sentence describing what is visible. Focus on the main subjects and actions. Keep it brief.
\end{tcolorbox}

\section{Additional Results}

\begin{table} [t]
\centering
\caption{Additional results on AVHBench and CMM. ACPO remains competitive with other baselines. AV Cap: audio-visual captioning (M: METEOR, C: CIDEr), VAL: vision-audio-language. PA: accuracy on yes-instances, HR: accuracy on no-instances. Best in bold.}
\label{tab:video}
\footnotesize
\setlength{\tabcolsep}{2.5pt}
\begin{tabular}{lcccccccccccc}
\toprule
& \multicolumn{6}{c}{\textbf{AVHBench}} & \multicolumn{6}{c}{\textbf{CMM}} \\
\cmidrule(lr){2-7}\cmidrule(lr){8-13}
& \multicolumn{4}{c}{Video Hallucination} & \multicolumn{2}{c}{AV Cap} & \multicolumn{3}{c}{VAL} & \multicolumn{3}{c}{Overrely Audio} \\
\cmidrule(lr){2-5}\cmidrule(lr){6-7}\cmidrule(lr){8-10}\cmidrule(lr){11-13}
\textbf{Method} & Prec$\uparrow$ & Rec$\uparrow$ & F1$\uparrow$ & Acc$\uparrow$ & M$\uparrow$ & C$\uparrow$ & PA$\uparrow$ & HR$\uparrow$ & Acc$\uparrow$ & PA$\uparrow$ & HR$\uparrow$ & Acc$\uparrow$ \\
\midrule
Base model \cite{videollama2} & 75.4 & 80.6 & 78.0 & 77.2 & 17.1 & 18.3 & 83.5 & 96.0 & 89.8 & 85.5 & 84.5 & \textbf{85.0} \\
SFT \cite{ouyang2022training} & 76.2 & 81.0 & \textbf{78.5} & 77.9 & 17.1 & 18.1 & 83.5 & 97.0 & 90.3 & \textbf{86.0} & 78.0 & 82.0 \\
DPO \cite{dpo} & 75.3 & \textbf{81.6} & 78.3 & 77.5 & \textbf{17.2} & \textbf{18.5} & \textbf{84.5} & 96.5 & \textbf{90.5} & \textbf{86.0} & 80.5 & 83.3 \\
OmniDPO \cite {omnidpo} & \textbf{79.3} & 77.5 & 78.4 & \textbf{78.7} & 16.8 & 16.1 & 79.5 & \textbf{98.5} & 89.0 & 81.0 & \textbf{85.5} & 83.3 \\
\rowcolor{green!18}Ours & 76.9 & 79.1 & 78.0 & 77.7 & \textbf{17.2} & 18.4 & 79.5 & 97.0 & 88.3 & 85.0 & 82.0 & 83.5 \\
\bottomrule
\end{tabular}
\end{table}

Table~\ref{tab:video} reports results on the remaining AVHBench and CMM tasks. We exclude vision-language tasks from CMM entirely, as performance on these tasks remains identical across all methods, since we modify only the audio projection layer. On AVHBench, ACPO matches or outperforms the base model on the video hallucination and audio-visual captioning tasks, confirming that audio-targeted training does not compromise visual understanding. On CMM overreliance on audio, ACPO exhibits the smallest drop relative to the base model among all preference-based methods. The modest decrease across all methods is expected. The base model's high score on this task partly reflects its tendency to ignore audio altogether, making it trivially robust to misleading audio. As methods improve audio grounding, the model naturally becomes more susceptible to audio interference, as it is actually attending to the audio signal.

\section{SFT on Audio-Swapped Data}

\begin{table}
\centering
\caption{Training with SFT on audio-swapped data, without preference optimization, hurts video hallucination while leaving audio hallucination roughly unchanged.}
\label{tab:sft-swaps}
\setlength{\tabcolsep}{8pt}
\begin{tabular}{lcccc}
\toprule
 & \multicolumn{2}{c}{Audio Halluc.} & \multicolumn{2}{c}{Video Halluc.} \\
\cmidrule(lr){2-3}\cmidrule(lr){4-5}
Model & Acc$\uparrow$ & F1$\uparrow$ & Acc$\uparrow$ & F1$\uparrow$ \\
\midrule
Base model~\cite{videollama2} & 74.0 & 77.3 & 77.2 & 78.0 \\
\quad + SFT (audio swaps)~\cite{ouyang2022training} & 74.0 & 74.9 & 64.3 & 66.7 \\
\bottomrule
\end{tabular}
\end{table}

To isolate whether ACPO's gains stem from its dual-axis preference objective or merely from exposure to audio-swapped captioning data, we train the base model with supervised fine-tuning (SFT) on audio-swapped captions using a standard cross-entropy loss. As Table~\ref{tab:sft-swaps} reports, SFT on the swapped data leaves audio hallucination essentially unchanged, while video hallucination accuracy drops from 77.2 to 64.3. Training on swapped captions alone thus does not improve the target audio task and degrades video grounding, confirming that ACPO's improvements come from its dual-axis preference objective rather than from the audio-swap augmentation alone.

\end{document}